\documentclass[pdflatex,sn-mathphys-num]{sn-jnl}

\usepackage{graphicx}%
\usepackage{multirow}%
\usepackage{amsmath,amssymb,amsfonts}%
\usepackage{amsthm}%
\usepackage{mathrsfs}%
\usepackage[title]{appendix}%
\usepackage{xcolor}%
\usepackage{textcomp}%
\usepackage{manyfoot}%
\usepackage{booktabs}%
\usepackage{algorithm}%
\usepackage{algorithmicx}%
\usepackage{algpseudocode}%
\usepackage{listings}%
\usepackage{adjustbox}

\usepackage{color,soul}

\theoremstyle{thmstyleone}%
\theoremstyle{thmstyletwo}%

\theoremstyle{thmstylethree}%

\begin{document}

\title[Article Title]{Aircraft Trajectory Dataset Augmentation in Latent Space}


\author[1]{\fnm{Seokbin} \sur{Yoon}}\email{sierra.bin@kau.ac.kr}

\author*[1]{\fnm{Keumjin} \sur{Lee}}\email{keumjin.lee@kau.ac.kr}

\affil[1]{\orgdiv{Department of Air Transport, Transportation, and Logistics}, \orgname{Korea Aerospace University}, \orgaddress{\street{76, Hanggongdaehak-ro, Deogyang-gu}, \city{Goyang}, \postcode{10540}, \state{Gyeonggi-do}, \country{Republic of Korea}}}


\abstract{Aircraft trajectory modeling plays a crucial role in air traffic management (ATM) and is important for various downstream tasks, including conflict detection and landing time prediction. Dataset augmentation by adding synthetically generated trajectory data is necessary to develop a more robust aircraft trajectory model and ensure that the trajectory dataset is sufficient and balanced. We propose a novel framework called ATRADA for aircraft trajectory dataset augmentation. In the proposed framework, a Transformer encoder learns the underlying patterns in the original trajectory dataset and converts each data point into a context vector in the learned latent space. The converted dataset is projected to reduced dimensions using principal component analysis (PCA), and a Gaussian mixture model (GMM) is applied to fit the probability distribution of the data points in the reduced-dimensional space. Finally, new samples are drawn from the fitted GMM, the dimension of the samples is reverted to the original dimension, and the samples are decoded with a multi-layer perceptron (MLP). Several experiments demonstrate that the framework effectively generates new, high-quality synthetic aircraft trajectory data, which were compared to the results of several baselines.}



\maketitle

\section{Introduction}\label{sec1}

As the demand for air transportation continues to increase around the world, the concept of air traffic operations has shifted to trajectory-based operation, which more accurately manages the four-dimensional trajectory  of aircraft in terms of latitude, longitude, altitude, and time. Many studies have developed data-driven aircraft trajectory models for various downstream applications such as conflict detection and landing time prediction~\cite{liu2011probabilistic,hong2015trajectory}.

One study suggested using a Gaussian mixture model (GMM) to learn statistical patterns of aircraft trajectories~\cite{barratt2018learning}, while another study used a long short-term memory (LSTM) network to learn aircraft trajectory patterns in the presence of various procedural constraints in airspace~\cite{shi20204}. Other research combined a convolutional neural network (CNN) with an LSTM network to better extract spatio-temporal features in trajectories~\cite{ma2020hybrid}. Recently, various efforts have been made to apply Transformer-based architectures for trajectory prediction~\cite{guo2022flightbert, guo2024flightbert++,tong2023long}.

\begin{figure}[t]			
	\centering
	\includegraphics[width=0.65\linewidth]{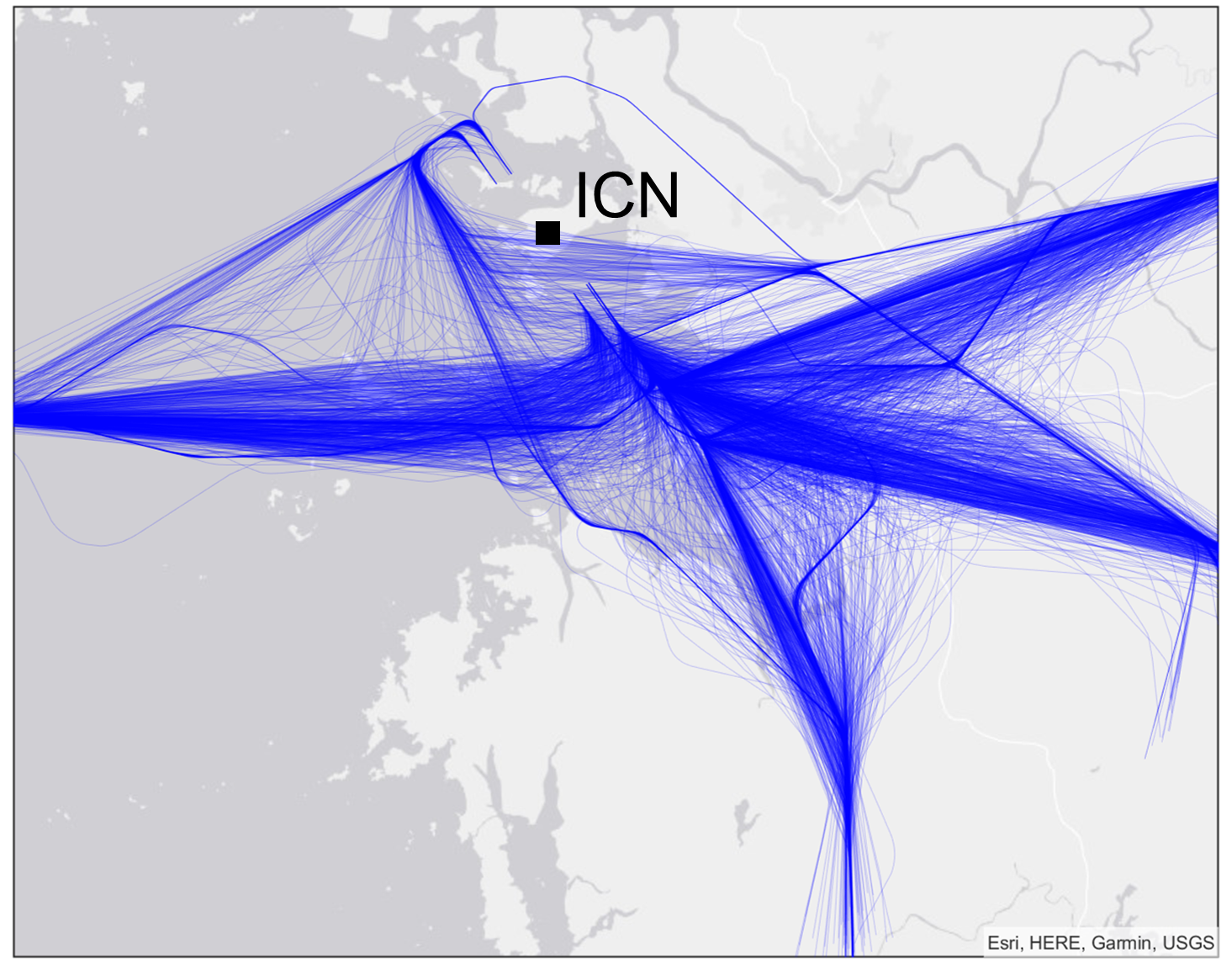}
    \caption{Actual aircraft trajectories at Incheon International Airport (ICN), South Korea, over a two-month period.}
    \label{trajectory}
 \end{figure}

\begin{figure}[t]			
	\centering
	\includegraphics[width=0.85\linewidth]{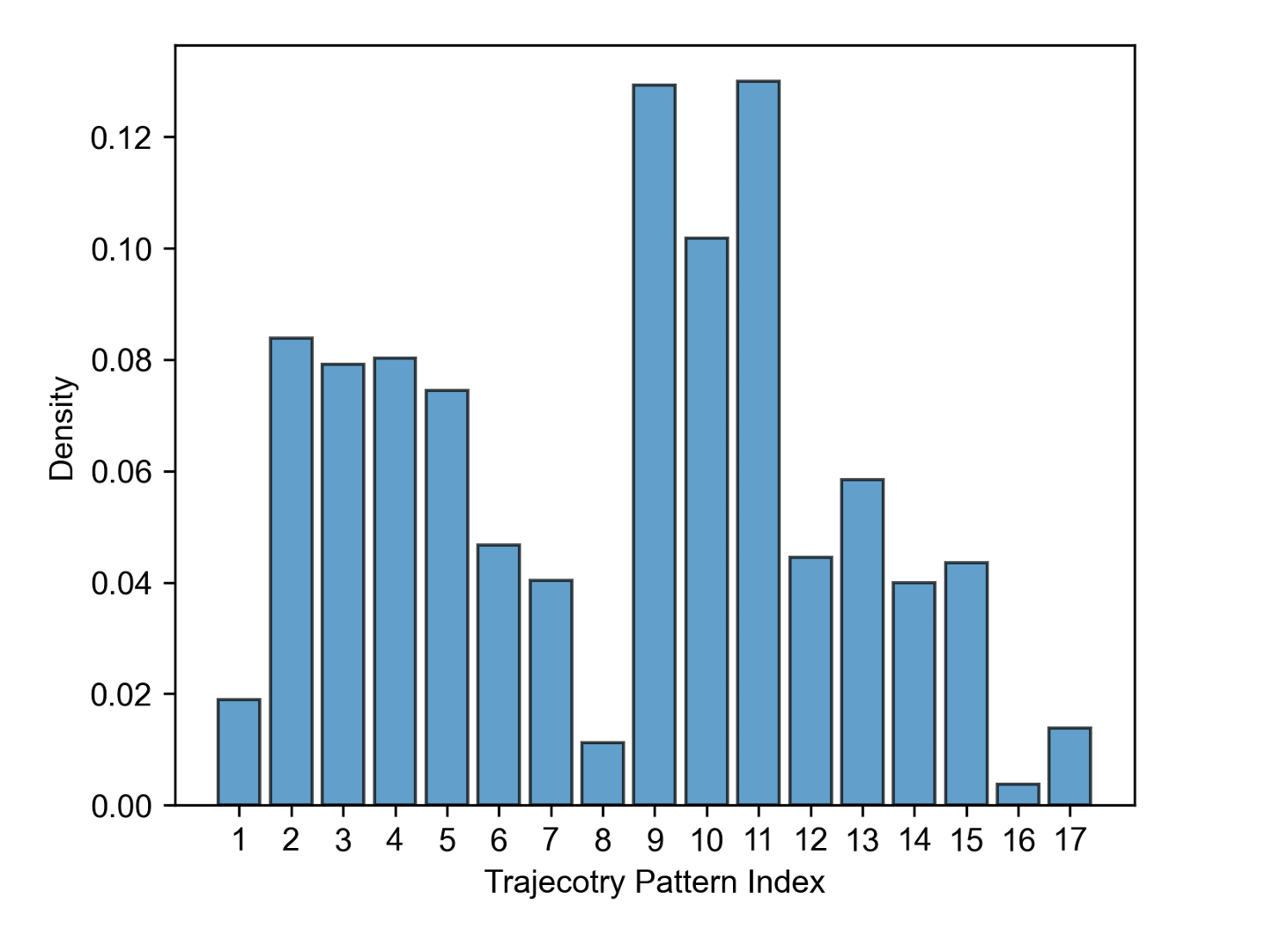}
    \vspace{-4mm} 
    \caption{Densities of trajectories in Figure~\ref{trajectory} for each trajectory pattern.}
    \label{trajectory_pattern}
 \end{figure}

Due to the complex interactions between aircraft, air traffic controllers (ATCs) often manually assign specific headings, altitudes, and speeds to aircraft to ensure safe and efficient operation. This process is known as radar vectoring and results in diverse patterns in aircraft trajectories, as shown in Figure~\ref{trajectory}. However, ATCs can have particular preferences regarding how to deviate aircraft from predefined flight paths, which can lead to imbalance among different trajectory patterns. Figure~\ref{trajectory_pattern} shows a histogram of different trajectory patterns for arriving aircraft at Inchon International Airport in South Korea, which illustrates the possibility of imbalance in the trajectory dataset. Such imbalance could deteriorate the performance of learning-based trajectory models and make them overfit major trajectory patterns. Therefore, it is beneficial to expand the dataset with synthetically generated data through dataset augmentation~\cite{yoon2023improving}.

Although a sufficient and balanced dataset is important for training data-driven aircraft trajectory models, very limited attention has been given to trajectory dataset augmentation. We propose a novel framework for augmentation that features explicit estimation of the underlying probability distribution of a trajectory dataset in latent space. This allows the augmented data to preserve the temporal dependency between trajectory points over time (aircraft dynamics) and the spatial flight patterns necessitated by various operational constraints in a specific airspace. The framework was validated using actual air traffic surveillance data. A series of experiments demonstrates that the synthetic trajectory data generated by the proposed method outperform those of other generative models in both discriminative and predictive tasks.

\section{Related Works}
\subsection{Time-series data generation} Several studies have applied generative adversarial networks (GANs)~\cite{goodfellow2014generativeadversarialnetworks} to generate synthetic time-series data. One study used LSTM networks as both a generator and discriminator. A noise vector was used as input, and time-series data were generated recursively~\cite{mogren2016crnngancontinuousrecurrentneural}. Another study applied a similar approach and generated more realistic time-series data using a recurrent neural network (RNN) with additional domain-specific inputs~\cite{esteban2017realvaluedmedicaltimeseries}. TimeGAN was proposed to better preserve the temporal dynamics in the time-series training data, and both reconstructive and adversarial objectives were jointly optimized~\cite{yoon2019time}. 

Variational autoencoders (VAEs) have been explored for time-series data generation. TimeVAE was introduced to enable user-defined distributions in the decoder and enrich the temporal properties of the generated data~\cite{desai2021timevaevariationalautoencodermultivariate}. The representation capabilities of transformers have led to their adoption in time-series generation tasks. For example, Museformer was designed with fine- and coarse-grained attention mechanisms for symbolic music generation~\cite{yu2022museformer}. Another study combined an adversarial autoencoder with a Transformer to generate time-series data by integrating local and global features in the data~\cite{srinivasan2022timeseriestransformergenerativeadversarial}.

\subsection{Aircraft trajectory generation} 
\begin{figure}[t]			
	\centering
	\includegraphics[width=0.6\linewidth]{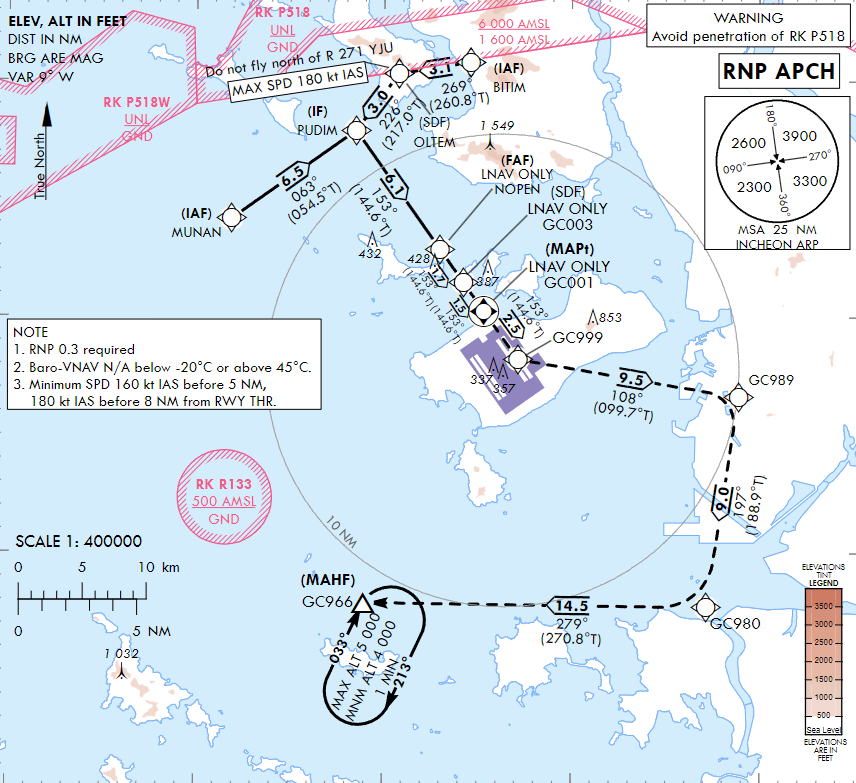}
    \caption{An example of operational constraints for arriving aircraft (Approach chart for ICN).}
    \label{RKSI}
\end{figure}

The generation of synthetic aircraft trajectories presents significant challenges because the trajectories need to capture both the temporal dynamics of aircraft movement and the spatial distribution while respecting various operational constraints within a specific airspace. For example, arriving aircraft must pass through specific three-dimensional fixes and maintain appropriate altitudes to ensure safe and efficient landing. One such critical point is the final approach fix (FAF), which marks the beginning of the final approach segment where pilots initiate their final descent for landing, as illustrated in Figure~\ref{RKSI}.

One study applied a GMM to fit the distribution of an aircraft trajectory dataset and generate synthetic trajectories by drawing samples from the fitted GMM~\cite{barratt2018learning}. Another study also applied a GMM and used a feature vector that consisted of deviations from a flight path rather than actual position measurements to better represent variability in trajectories within specific airspace~\cite{jung2023inferringtrafficmodelsterminal}. The synthetic minority over-sampling technique (SMOTE)~\cite{chawla2002smote} was used to augment aircraft trajectory data in minor trajectory patterns to improve the accuracy of trajectory prediction~\cite{yoon2023improving}. Another study combined a temporal convolutional network (TCN) and a VAE for aircraft trajectory generation~\cite{krauth2023deep}.

\section{Methodology}
The proposed framework consists of three parts. First, the original trajectory dataset is transformed into a learned latent space using a sequence autoencoder trained to reconstruct the input trajectory in its output. Various architectures, including a sequence to sequence (Seq2Seq) LSTM network~\cite{sutskever2014sequence}, can be used as an autoencoder. However, we chose a Transformer-based architecture due to its better representation capability for long sequences~\cite{vaswani2017attention}. Second, once the original trajectory dataset is transformed into a set of context vectors in the latent space, the probability distribution of the vectors is fitted by a GMM. Lastly, new samples are drawn from the fitted GMM and decoded in the same manner as in the trajectory reconstruction process in the first step.

\subsection{Background}
\subsubsection{Self-attention mechanism} The self-attention mechanism is the heart of the Transformer~\cite{vaswani2017attention} and mimics human cognitive processes by selectively focusing on specific elements within an input sequence. The attention mechanism is defined as:

\begin{equation}
\text{Attention}(\mathbf{Q},\mathbf{K},\mathbf{V})= \text{softmax}(\frac{\mathbf{Q}\mathbf{K}^T}{\sqrt{d_k}})\mathbf{V}
\end{equation}

where $\mathbf{Q}$, $\mathbf{K}$, and $\mathbf{V}$ represent queries, keys, and values, respectively, and $d_k$ represents the dimension of the keys. The queries $\mathbf{Q}$ interact with the keys $\mathbf{K}$ through a dot-product operation that quantifies the relevance of each key to each query. This interaction produces attention weights that are normalized using the softmax function, which are applied to the values $\mathbf{V}$ to contextualize the input sequence. In practice, multi-head attention is applied to provide a more diverse representation of the input sequence.

\subsubsection{Gaussian mixture model}
The GMM is a probabilistic model that represents the probability distribution of a dataset as a mixture of multiple Gaussian distributions, each representing a distinct cluster of data points. The GMM is defined as:
\begin{equation}
p(\mathbf{t}) = \sum_{i=1}^K \pi_i \mathcal{N}(\mathbf{t} ; \mu_i, \Sigma_i) 
\end{equation}
where $\mathbf{t}$ is a $C$-dimensional vector with a probability distribution comprising the weighted sum of $K$ Gaussian distributions. $\pi_i$ denotes the mixture weight of the $i^{\text{th}}$ Gaussian component, which has a constraint that the sum of all mixture weights must be equal to 1. The parameters $\{\pi_i, \mu_i, \Sigma_i\}$ of each $i^{\text{th}}$ Gaussian distribution are estimated using the expectation-maximization (EM) algorithm to maximize the likelihood of the observed data.

\begin{figure*}[t!]			
	\centering
	\includegraphics[width=0.97\textwidth]{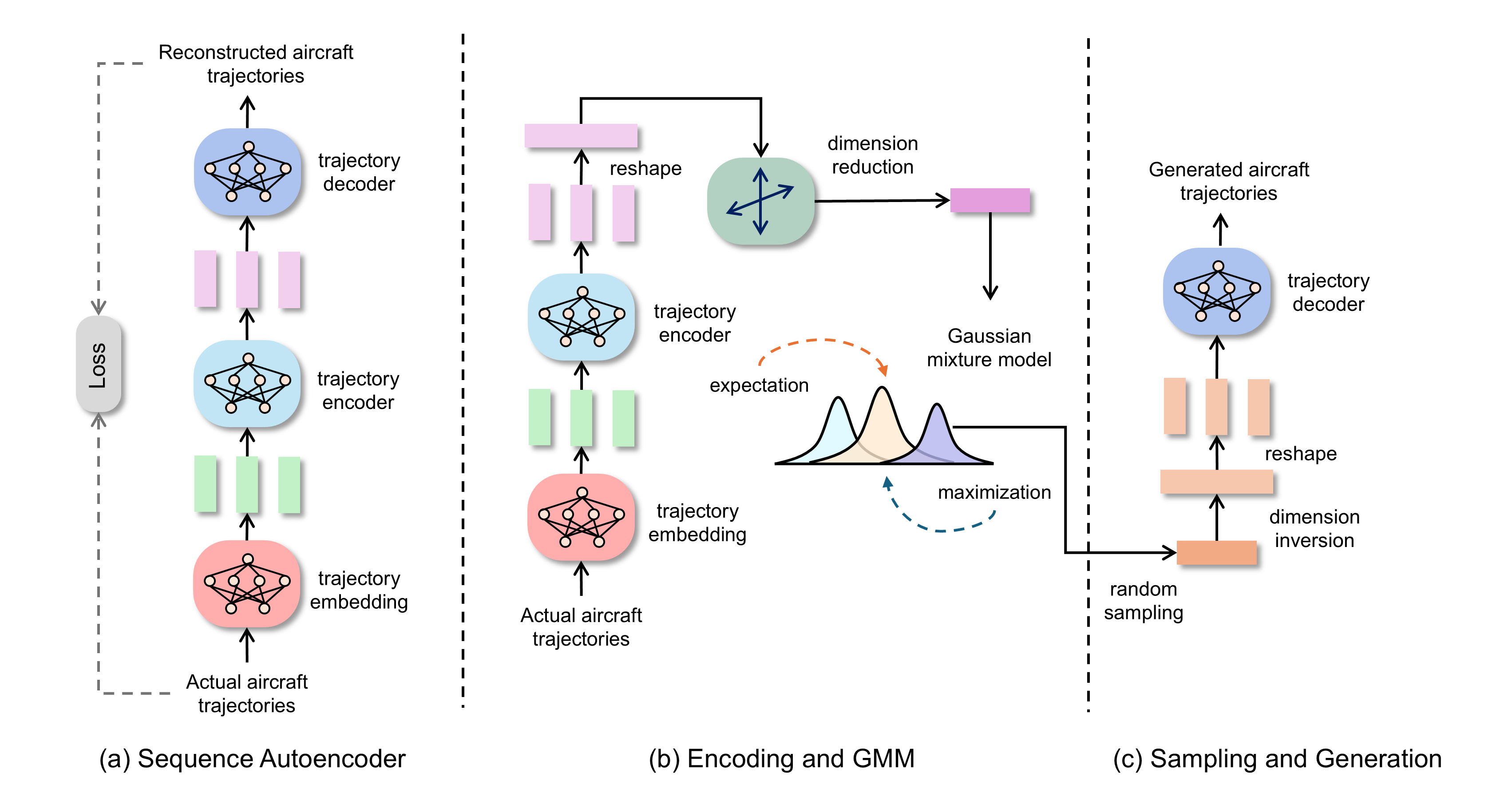}
    
    \caption{The proposed framework for aircraft trajectory dataset augmentation. (a) A sequence autoencoder learns latent space representations in an unsupervised way. (b) Aircraft trajectories are encoded, and GMM is fitted to the latent vectors. (c) Samples are randomly drawn from the fitted GMM and decoded for generation.}
    \label{overview}
\end{figure*}

\subsection{Proposed framework}
As illustrated in Figure~\ref{overview}, the proposed framework consists of two main components: an autoencoder and a GMM. The autoencoder is trained to reconstruct the input trajectory and then used to convert the trajectory dataset from its original space into context vectors in the latent space. The decoder part of the autoencoder synthesizes new trajectory data from samples drawn from the GMM, which fits the probability distribution of the dataset in the latent space. Importantly, since the GMM training occurs after the autoencoder is fully trained, the generation results can be optimized without retraining the autoencoder by adjusting the associated hyperparameters: the number of principal components for principal component analysis (PCA) and the number of Gaussian distributions in the GMM.

\subsubsection{Sequence autoencoder} For the sequence autoencoder, we use a Transformer-based encoder with a self-attention mechanism and a multi-layer perceptron (MLP) as a decoder. As illustrated in Figure~\ref{overview}(a), we first embed the aircraft trajectory $\{\mathbf{x}^i_{1:S}\}^B_{i=1} \in \mathbb{R}^{B\times F \times S}$ into the embedding vectors with a hidden dimension $D$, where $B$ is the mini-batch size, $F$ is the number of original features (latitude, longitude, and altitude), and $S$ is the maximum length of aircraft trajectory sequences. The embedded trajectory sequence $\{\mathbf{x_e}^i_{1:S}\}^B_{i=1} \in \mathbb{R}^{B\times D \times S}$ is then fed into the trajectory encoder. 

The trajectory encoder uses the full self-attention mechanism, as proposed in the original Transformer model~\cite{vaswani2017attention}. This choice is based on the equal importance of the long-term relationships between trajectory points at different timestamps to that of the short-term relationships in air traffic operations. For example, the trajectory points in the early part of a sequence contain information about the entry point into the airspace and significantly influence the trajectory points at much later time steps due to the use of standard terminal arrival route (STAR). Finally, the latent vectors $\{\mathbf{z}^i_{1:S}\}^B_{i=1} \in \mathbb{R}^{B\times D \times S}$ from the trajectory encoder are passed into the decoder, where the aircraft trajectory is reconstructed as $\{\mathbf{\tilde{x}}^i_{1:S}\}^B_{i=1} \in \mathbb{R}^{B\times F \times S}$. We measure the difference between the actual trajectories and the reconstructed trajectories using an L2 loss function, $\frac{1}{B}\Sigma_{i=1}^B\left\| \mathbf{x}^i - \mathbf{\tilde{x}}^i \right\|_2^2$. Notably, the use of the MLP in the trajectory decoder enables non-autoregressive trajectory generation, which significantly reduces the computation time and minimizes the risk of error accumulation that typically occurs in other autoregressive times-series generative models.

\subsubsection{Encoding and GMM} Once the autoencoder is fully trained, we encode the aircraft trajectory sequences $\{\mathbf{x}^j_{1:S}\}^N_{j=1} \in \mathbb{R}^{N\times F\times S}$ from the original dataset into latent vectors $\{\mathbf{z}^j_{1:S}\}^N_{j=1} \in \mathbb{R}^{N\times D \times S}$, where $N$ is the total number of trajectories in the dataset. We then transpose the latent vector at each time step and concatenate them to form a row vector $\overrightarrow{\mathbf{z}_j}$ for each trajectory sequence: 

\begin{equation}
    \overrightarrow{\mathbf{z}}^j = \{\mathbf{z}^T_{j_{1}} \oplus \mathbf{z}^T_{j_{2}} \oplus \cdots \oplus \mathbf{z}^T_{j_{S}}\}
\end{equation}
where $\overrightarrow{\mathbf{z}}^j \in \mathbb{R}^{1 \times (D \cdot S)}$, and $\oplus$ denotes concatenation. $\overrightarrow{\mathbf{z}}^j$ is then stacked row-wise, which results in a matrix $\mathbf{M}\in \mathbb{R}^{N\times (D \cdot S)}$. PCA is applied to this matrix, which results in a matrix $\mathbf{O} \in \mathbb{R}^{N\times P}$. Through PCA, the latent vectors $\overrightarrow{\mathbf{z}}^j$ are projected into a low-dimensional space $(1 \leq P < (D \cdot S))$, where the principal feature vectors are orthogonal to each other. This step helps to manage redundancy and noise in the dataset and facilitates the training of the GMM, which fits the matrix $\mathbf{O}$ as a mixture of multiple Gaussian distributions $p(\mathbf{y}) = \sum_{i=1}^K \pi_i \mathcal{N}(\mathbf{y} ; \mu_i, \Sigma_i)$.

\subsubsection{Sampling and generation} To augment a dataset, new latent vectors $\{\mathbf{y}^q\}^M_{q=1} \in \mathbb{R}^{M\times P}$ are sampled from the fitted GMM, where $M$ is the number of generated samples. The vectors can then be reverted to the original dimension $(\mathbb{R}^{M\times P} \rightarrow \mathbb{R}^{M \times (D \cdot S)})$ and reshaped $(\mathbb{R}^{M\times (D \cdot S)} \rightarrow \mathbb{R}^{M\times D \times S})$ for decoding. New aircraft trajectory data $\{\tilde{\mathbf{y}}^q_{1:S}\}^M_{q=1} \in \mathbb{R}^{M\times F\times S}$ are then generated as outputs of the decoder. Importantly, the new samples $\{\mathbf{y}^q\}^M_{q=1}$ can still be effectively decoded, as the multivariate Gaussian distribution $p(\mathbf{y})$ provides proper representations of the original dataset. 

\section{Numerical Experiments}
The proposed framework was implemented with the following software and hardware: Python 3.9.16, PyTorch 2.0.1, CUDA 11.7, an Intel Core i7-12700F CPU, 32 GB of RAM, and an NVIDIA GeForce RTX 3060 GPU.

\subsection{Dataset description} We utilized trajectories of arriving aircraft in the terminal airspace (70-NM radius) at Incheon International Airport. The dataset was acquired from FlightRadar24\footnote{\url{https://www.flightradar24.com}}, which collects aircraft trajectory data via the automatic dependent surveillance–broadcast (ADS-B) system. The dataset covers the period of January 2022 to June 2022 and comprises sequential records of the state of each aircraft. The data include details such as the time, aircraft type, geographical position (latitude, longitude, and altitude), and other pertinent information. The ADS-B trajectory data have a non-uniform sampling rate, so we resampled the data using a piecewise cubic Hermite interpolating polynomial (PCHIP)~\cite{fritsch1980monotone} to make the rate uniform ($\Delta t = 6 \ \text{sec}$). Table~\ref{dataset} provides a description of the ADS-B trajectory dataset used in this work.

\begin{table}[h!]
\centering
\large
\caption{ADS-B trajectory dataset description.}
\begin{tabular}{l|l}
\toprule
\textbf{Attribute} & \textbf{Detail}  \\
\midrule
Number of Seq& 7,712  \\
Dimension & 3   \\
Avg Seq Len & 264 (1,584 sec) \\
Max Seq Len & 271 (1,626 sec)\\
Features & Latitude, Longitude, Altitude\\
\bottomrule
\end{tabular}
\label{dataset}
\end{table}

\subsection{Model configurations} We set the number of encoder layers to $L=3$, the hidden dimension to $D=128$, the number of heads to $h=4$, the feedforward dimension to $D_{ff}=512$, and the dropout probability to $p=0.2$. For training, we used the ADAM optimizer~\cite{kingma2017adammethodstochasticoptimization} with an initial learning rate of $10^{-4}$. The number of principal components $P$ for PCA was set to 22, and the number of Gaussian components $K$ for GMM was set to 32.

\begin{figure}[t]			
	\centering
	\includegraphics[width=0.85\linewidth]{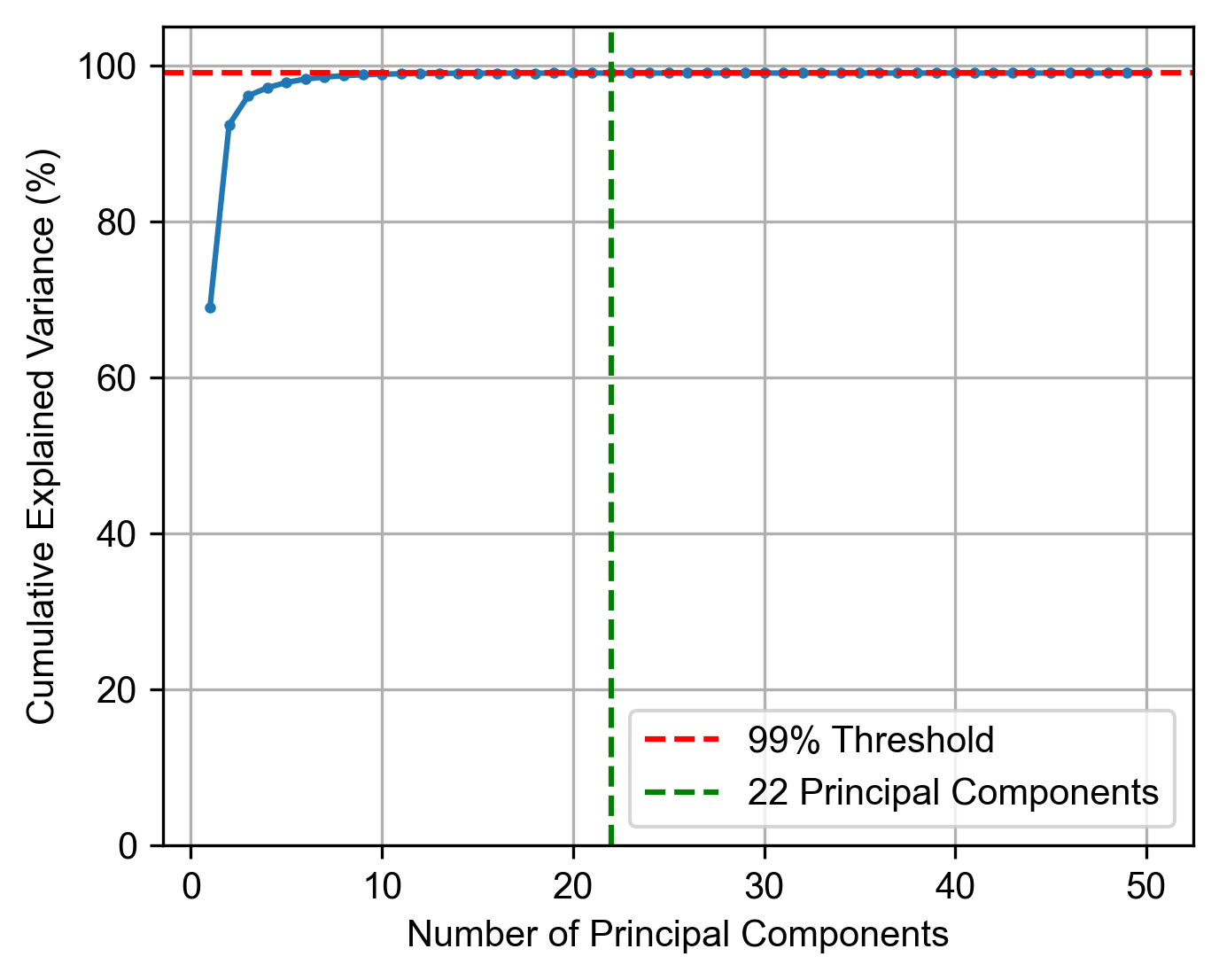}
    \caption{Sensitivity analysis of cumulative explained variance with respect to the number of principal components in PCA.}
    \label{pca}
 \end{figure}

\begin{figure}[t]			
	\centering
	\includegraphics[width=0.85\linewidth]{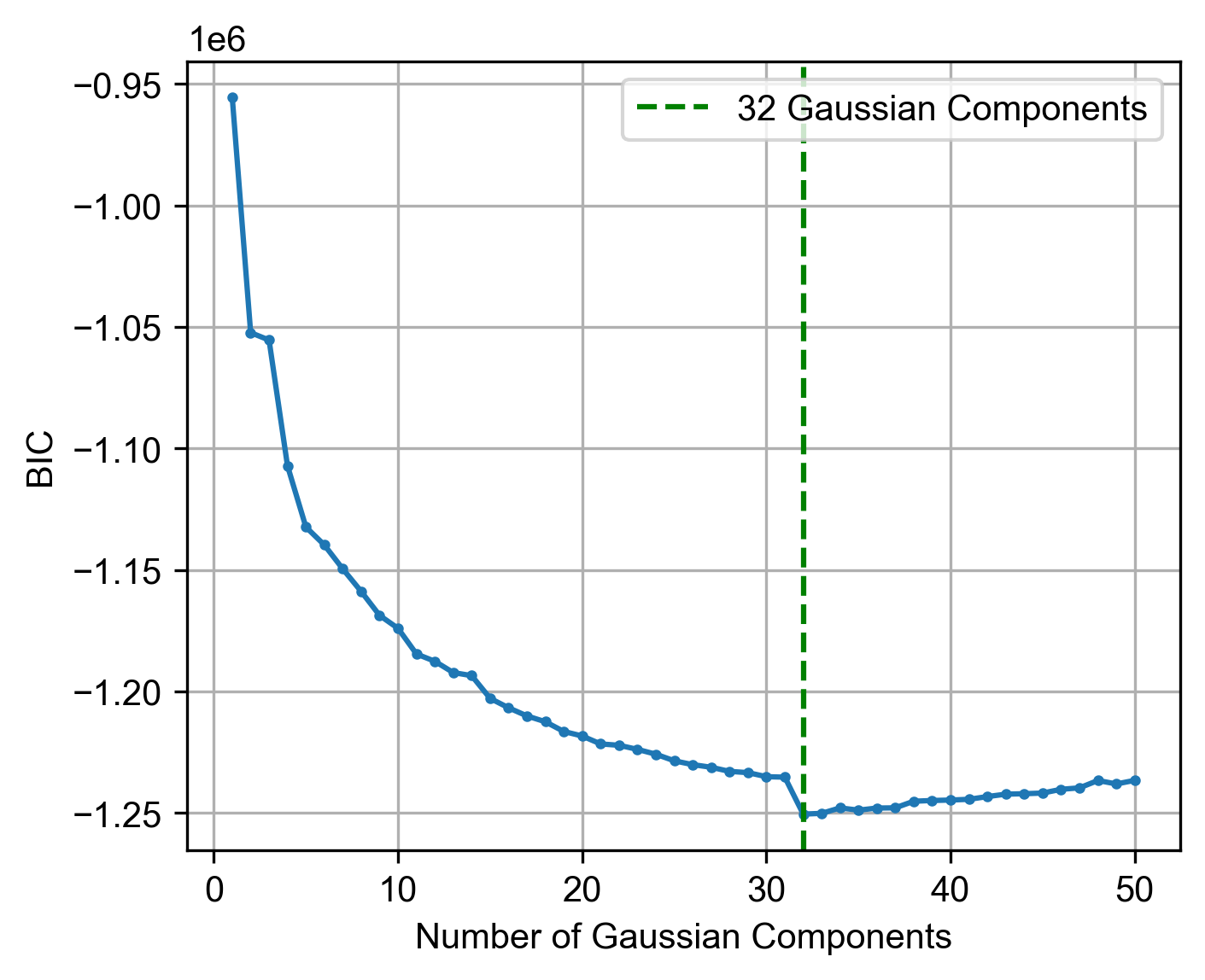}
    \caption{Sensitivity analysis of BIC with respect to the number of Gaussian components of GMM.}
    \label{gmm}
 \end{figure}

To determine $P$, we tested different values from 1 to 50 and selected 22 as the minimum number of components required to explain 99\% of the cumulative variance. The scree plot for the PCA is presented in Figure~\ref{pca}. For the GMM, we evaluated the Bayesian information criterion (BIC)~\cite{schwarz1978estimating} for values of $K$ ranging from 1 to 50, and the optimal value was chosen as $K=32$, which achieved the lowest BIC, thus balancing the model likelihood and complexity. The BIC plot for different values of $K$ is shown in Figure~\ref{gmm}. 

\subsection{Comparison baselines} We compared the proposed framework with several competitive baselines: TimeGAN~\cite{yoon2019time}, TimeVAE~\cite{desai2021timevaevariationalautoencodermultivariate}, SMOTE-Interpolation (SMOTE-I), SMOTE-Extrapolation (SMOTE-E)~\cite{devries2017datasetaugmentationfeaturespace}, GMM~\cite{barratt2018learning} and TCN-VAE~\cite{krauth2023deep}. We used the following publicly available source code for the implementation of the baselines (TimeGAN, TimeVAE, TCN-VAE and GMM). The source code for ATRADA will also be made publicly available.

\begin{itemize}
    \item TimeGAN~\cite{yoon2019time}: \url{https://github.com/jsyoon0823/TimeGAN}
    \item TimeVAE~\cite{desai2021timevaevariationalautoencodermultivariate}: \url{https://github.com/abudesai/timeVAE}
    \item TCN-VAE~\cite{krauth2023deep}:  \url{https://github.com/kruuZHAW/deep-traffic-generation-paper}
    \item GMM~\cite{barratt2018learning}: \url{https://github.com/sisl/terminal-airspace-models}
\end{itemize}

For TimeGAN, the hidden dimensions of its three gated recurrent unit (GRU) layers were set to four times the size of the input features, which is consistent with the implementation described in the original TimeGAN paper~\cite{yoon2019time}. The hidden dimensions of TimeVAE and TCN-VAE were configured to match the hidden dimension used in ATRADA. For both SMOTE-I and SMOTE-E~\cite{devries2017datasetaugmentationfeaturespace}, we set the number of nearest neighbors to 10, and the degrees of interpolation and extrapolation were set as 0.5. Given that SMOTE-I and SMOTE-E were tailored to generate 10 times more trajectories than the original data, we only used every tenth generated sample for each trajectory data point. We used the same number of Gaussian components for the implementation of the GMM as in ATRADA. Furthermore, to address the issue of noisy covariance matrices caused by high dimensionality, we used the singular value decomposition technique described in the original paper~\cite{barratt2018learning}.

\subsection{Quantitative evaluation} We performed discriminative and predictive tasks to quantitatively evaluate the quality of the generated aircraft trajectory data. For the discriminative task, we trained a Transformer-based classifier to distinguish between actual and generated trajectories. For the classification, we appended a classification (CLS) token to the input sequence, which is the approach used for bidirectional encoder representations from Transformers (BERT) and Vision Transformer (ViT)~\cite{devlin2019bertpretrainingdeepbidirectional,dosovitskiy2021imageworth16x16words}. This CLS token is used for classifying a given trajectory as either \textit{real} or \textit{generated}. We then computed the discriminative score as $\left\vert \frac{1}{2} - \text{accuracy} \right\vert$ (DS-Classifier). A score closer to zero indicates that the classifier is unable to distinguish between actual and generated trajectories and performs no better than random guessing. 

We also conducted an ATC Turing test by engaging four professional ATCs who specialize in the airspace of ICN. They were asked to distinguish between real and generated trajectories among 10 trajectory snapshots, which displayed both a horizontal view and a vertical profile of the trajectory. Half of these snapshots portrayed real trajectories, while the remaining five displayed generated trajectories. The discriminative score for the ATC Turing test (DS-ATC) was computed similarly to DS-Classifier.

Next, we trained a Transformer-based predictor on the generated aircraft trajectory dataset and evaluated its performance on an actual aircraft trajectory dataset, which is known as the “train on synthetic, test on real” (TSTR) approach. The predictor was designed to predict a future trajectory sequence over a time horizon of 120 seconds given a past trajectory sequence over a time horizon of 120 seconds. For training the predictor, we created pairs of past and future trajectory sequences spanning 20 points each (120 seconds) by sliding a window with a size of one time step from the generated trajectories. We used the mean absolute error (MAE) to assess the prediction accuracy (PS). Trajectory prediction is one of the major downstream applications of trajectory dataset augmentation, so it is important to validate the usefulness of the generated trajectories based on prediction accuracy.

We also directly evaluated the quality of the generated trajectories and assessed their statistical similarity to the actual data by calculating the Kullback–Leibler divergence (KLD) between the actual and synthetic trajectory distributions. KLD quantifies the discrepancy between the probability distribution of the actual data and that of the generated trajectories and is defined as follows:
\begin{equation}
   \text{KLD}(P \parallel Q) = \sum_{i} P(x_i) \log \frac{P(x_i)}{Q(x_i)}
\end{equation}


where $P$ and $Q$ denote the probability distributions of the actual and generated trajectories, respectively.

\subsection{Qualitative evaluation} We applied t-distributed stochastic neighbor embedding (t-SNE)~\cite{van2008visualizing} to the datasets of actual and generated aircraft trajectories to visualize how well they match each other. For deeper analysis, we also applied it to the first and second derivatives of aircraft trajectories (velocity and acceleration) to assess whether the generated trajectories adequately capture the dynamic behavior of aircraft over time. We visualized the top view (latitude and longitude) of the generated trajectories to evaluate how well the generated trajectories satisfy various operational constraints, including the designated entry points to the airspace and the FAF that aircraft must pass through to prepare for landing.

\section{Results}
\subsection{Quantitative evaluation} Table~\ref{tab:model_comparison} presents the experimental results of the quantitative evaluation. Each column shows the scores of ATRADA and the baseline models for each evaluation metric. DS-Classifier, DS-ATC, and KLD require comparison between actual and generated trajectory data, so reporting these scores for the original dataset is not applicable, and they were omitted from the table.

The relative improvements (RIs) of ATRADA over the second-best baseline models are provided at the bottom of the table. RI was calculated as $\left\vert s_2 - s_1 \right\vert / s_2$ , where $s_1$ denotes the best score achieved by the proposed method (indicating the lowest error), and $s_2$ is the second-best score among the baselines. In other words, the RI represents the relative error reduction of ATRADA compared to the second-best baseline for each evaluation metric.

ATRADA outperformed all baseline models and achieved RIs of 43.4\%, 25.0\%, 36.0\%, and 17.1\% across the four evaluation metrics compared to TimeVAE, SMOTE-I, GMM, and GMM, respectively. This result indicates that ATRADA generates trajectories that are more indistinguishable from actual trajectories and more effectively preserves their predictive characteristics. Furthermore, the KLD results confirm that ATRADA generates trajectories that are statistically closer to the actual trajectories.

TimeVAE and GMM also performed well. Specifically, the trajectories generated by TimeVAE achieve the second-best score for DS-Classifier metric and the third-best score for DS-ATC metric. However, they tended to be less effective for predictive tasks compared to ATRADA and GMM. On the other hand, GMM generated trajectories that better preserve the predictive characteristics of the actual trajectories (lower PS) than TimeVAE, but its discriminative scores were not as strong as those of ATRADA and TimeVAE. SMOTE-I generates trajectories by interpolating two latent vectors and produced trajectories that are similar to actual trajectories, as indicated by strong scores for the DS-Classifier and DS-ATC metrics. However, these generated trajectories were less useful for predictive tasks. Conversely, the opposite trend occurred with SMOTE-E, which extrapolates two latent vectors to generate synthetic trajectories.

\begin{table}[t!]
\centering
\large
\caption{Experimental results for four quantitative evaluations. Bold and underlined values indicate the best and second-best performance, respectively.}
\begin{tabular}{c|c|c|c|c}
\toprule
\multirow{2}{*}{Model} & DS-Classifier & DS-ATC & PS & KLD  \\
       & \multicolumn{4}{c}{(Lower the Better)} \\   
\midrule
ATRADA (Ours) & \textbf{0.013} & \textbf{0.075} & \textbf{0.016} & \textbf{44.001} \\
\midrule
TimeGAN & 0.414 & 0.350 & 0.084 & 2706.021 \\
TimeVAE & \underline{0.023} & 0.125 & 0.028  & 75.472 \\
SMOTE-I & 0.027 & \underline{0.100} & 0.056 & 60.525 \\
SMOTE-E  & 0.176 & 0.375 & 0.031 & 762.927  \\
GMM & 0.057 & 0.300 & \underline{0.025}  & \underline{53.061} \\
TCN-VAE & 0.036 & 0.125 & 0.037  & 673.747 \\
\midrule
Original & - & - & 0.009 & - \\
\midrule

RI & \textbf{43.4\%} & \textbf{25.0\%} & \textbf{36.0\%}  & \textbf{17.1\%}\\

\bottomrule
\end{tabular}
\label{tab:model_comparison}
\end{table}

\subsection{Qualitative evaluation} 
\begin{figure*}[t!]			
	\centering
	\includegraphics[width=.95\textwidth]{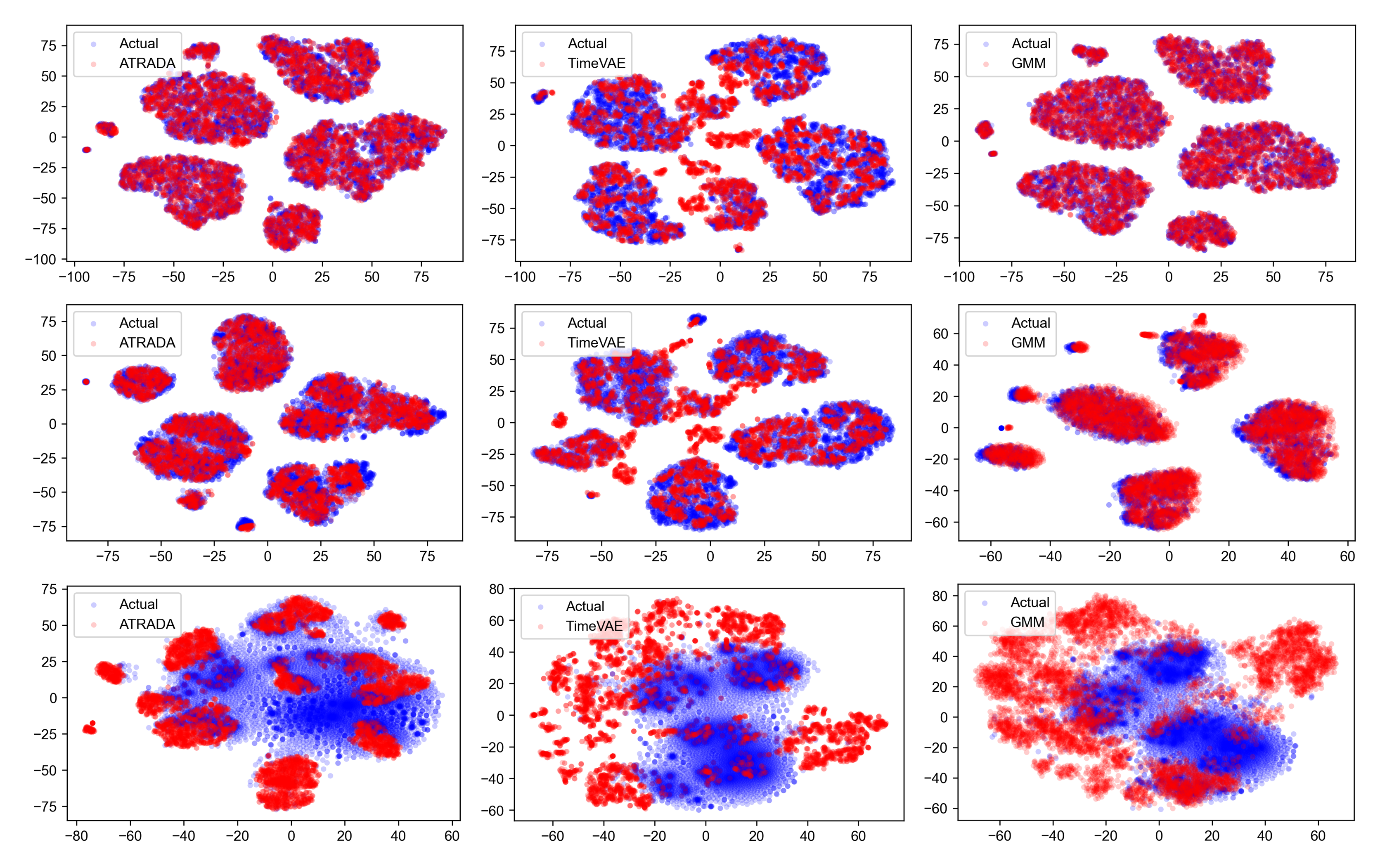}
    \caption{t-SNE visualizations for ATRADA (left), TimeVAE (center), and GMM (right). The t-SNEs for the generated trajectories’ position are in the 1$^{\text{st}}$ row, velocity (first derivative) in the 2$^{\text{nd}}$ row, and acceleration (second derivative) in the 3$^{\text{rd}}$ row. Blue indicates the actual trajectories, and red indicates the generated trajectories.}
    \label{tsne}
 \end{figure*}

Figure~\ref{tsne} presents the t-SNE visualizations for the position ($1^{\text{st}}$ row), velocity ($2^{\text{nd}}$ row), acceleration ($3^{\text{rd}}$ row) for the generated trajectories by ATRADA, TimeVAE, and GMM, which produced good performance in the quantitative evaluation. The t-SNE visualizations for other baselines can be found in Appendix~\ref{appendix_figure}. The trajectories generated by ATRADA exhibit strong overlap with the actual trajectories across all three dimensions. In contrast, a significant portion of the trajectories from TimeVAE and GMM fall out of the distribution for acceleration, which indicates that they do not adequately capture the dynamic behavior of aircraft. 

\begin{figure*}[t!]			
	\centering
	\includegraphics[width=.9\textwidth]{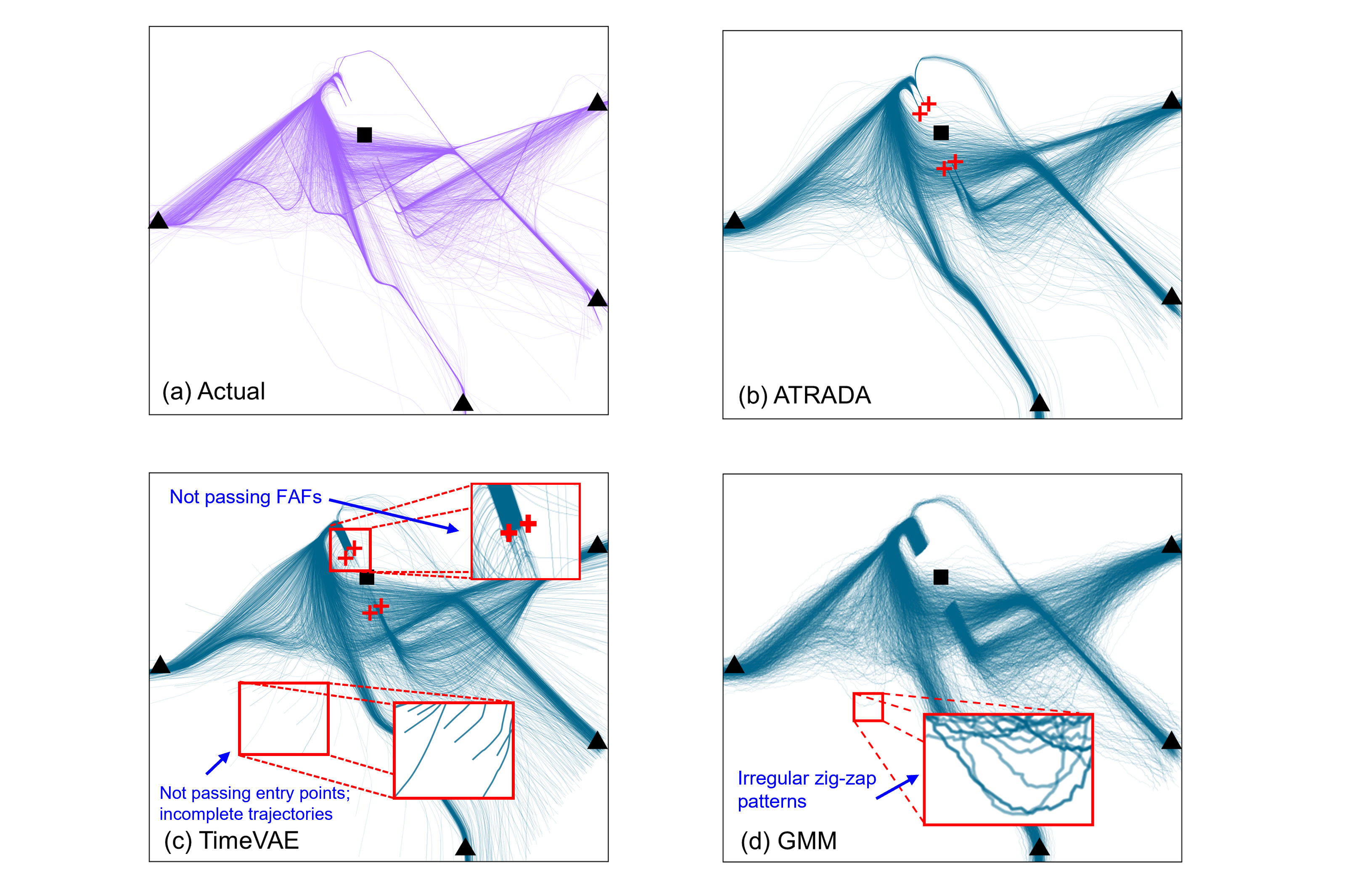}
    \caption{Top view of generated trajectories by ATRADA, TimeVAE, and GMM.}
    \label{topview}
 \end{figure*}

Notably, TimeVAE also generated some out-of-distribution samples for position and velocity, which correspond to synthetic trajectories with arbitrary entry points into the airspace, as shown in Figure~\ref{topview}(c). Note that during the process of projecting each trajectory into a 2-demensional space using t-SNE, the x- and y-axes are determined by the distribution of the combined dataset (i.e., the actual trajectories together with the synthetic trajectories generated by each method). As a result, although the same set of actual trajectories is used across all three methods (ATRADA, TimeVAE, and GMM), their visual shapes appear different in each subfigure of Figure~\ref{tsne}.

Figure~\ref{topview} presents top views of actual and generated trajectories, with entry points and the airport marked by black triangles and squares, respectively. The FAFs for different runway directions are also shown by red symbols (“+”). ATRADA and TimeVAE both generated smooth aircraft trajectories in the top view. However, TimeVAE generated trajectories that spread widely across the airspace and often failed to pass through the designated airspace entry points or the FAFs, as highlighted by the red boxes in Figure~\ref{topview}(c). Trajectories that do not pass through the FAFs may imply landings between runways or outside valid runway boundaries, which cannot occur in actual operations.

Similarly, GMM generates trajectories that often fail to pass through the FAFs and exhibit highly irregular zig-zag patterns that are unrealistic for aircraft movements, as highlighted by the red boxes in Figure~\ref{topview}(d). In contrast, ATRADA consistently generated trajectories that precisely pass through these points, which demonstrates their realism and their potential utility for downstream tasks.

\section{Discussion}
TimeGAN did not perform well in the task of aircraft trajectory generation despite being considered a state-of-the-art method for time-series data generation. We conjecture that this resulted from its implicit likelihood training. Specifically, the generator of TimeGAN is designed to deceive the discriminator rather than accurately learn the probability distribution of the dataset, so the trajectories generated may fail to satisfy critical spatial constraints, such as designated entry points or STARs in the airspace. 

In contrast, the models with explicit likelihood training (ATRADA, TimeVAE, TCN-VAE, and GMM) performed better in generating trajectories that align with the spatial distribution of actual trajectories. However, TimeVAE and TCN-VAE still showed limited capability in accurately capturing the distribution of actual trajectories because they learn the dataset’s distribution indirectly through reconstruction and regularization objectives. Conversely, ATRADA directly estimates the parameters of Gaussian distributions from the actual dataset, which can lead to more consistent and reliable trajectory generation.

The idea of using the context vectors in the latent space rather than the original variables has shown statistical advantages in many prior works~\cite{bengio2012bettermixingdeeprepresentations,devries2017datasetaugmentationfeaturespace}. The results of this work are also consistent with these findings. Working with GMM in the latent space rather than in the original input space allows us to generate synthetic trajectories that are not only more similar to real trajectories but also more beneficial for the downstream tasks, such as trajectory prediction.

\section{Conclusion}\label{sec13}
We have proposed a novel framework for aircraft trajectory dataset augmentation that leverages the representational capabilities of the Transformer as a sequence autoencoder and the probabilistic nature of the GMM as a generator. The effectiveness of ATRADA was validated using actual air traffic surveillance data. The results demonstrated that it generates more realistic aircraft trajectories that respect the operational constraints of real-world airspace. Given the highly constrained nature of air traffic operations due to safety concerns, the ability of ATRADA to generate trajectories that adhere to both aerodynamic and operational constraints is particularly advantageous. In summary, ATRADA exhibits better performance in generating realistic aircraft trajectories, and could be beneficial for various downstream tasks, such as trajectory prediction and air traffic simulation.

\backmatter

\section*{Acknowledgements}
This work is supported by the Korea Agency for Infrastructure Technology Advancement (KAIA) grant funded by the Ministry of Land, Infrastructure and Transport (Grant RS-2022-00156364).

\begin{appendices}

\section{Additional Visualizations}\label{appendix_figure}
\begin{figure*}[t!]			
	\centering
	\includegraphics[width=0.89\textwidth]{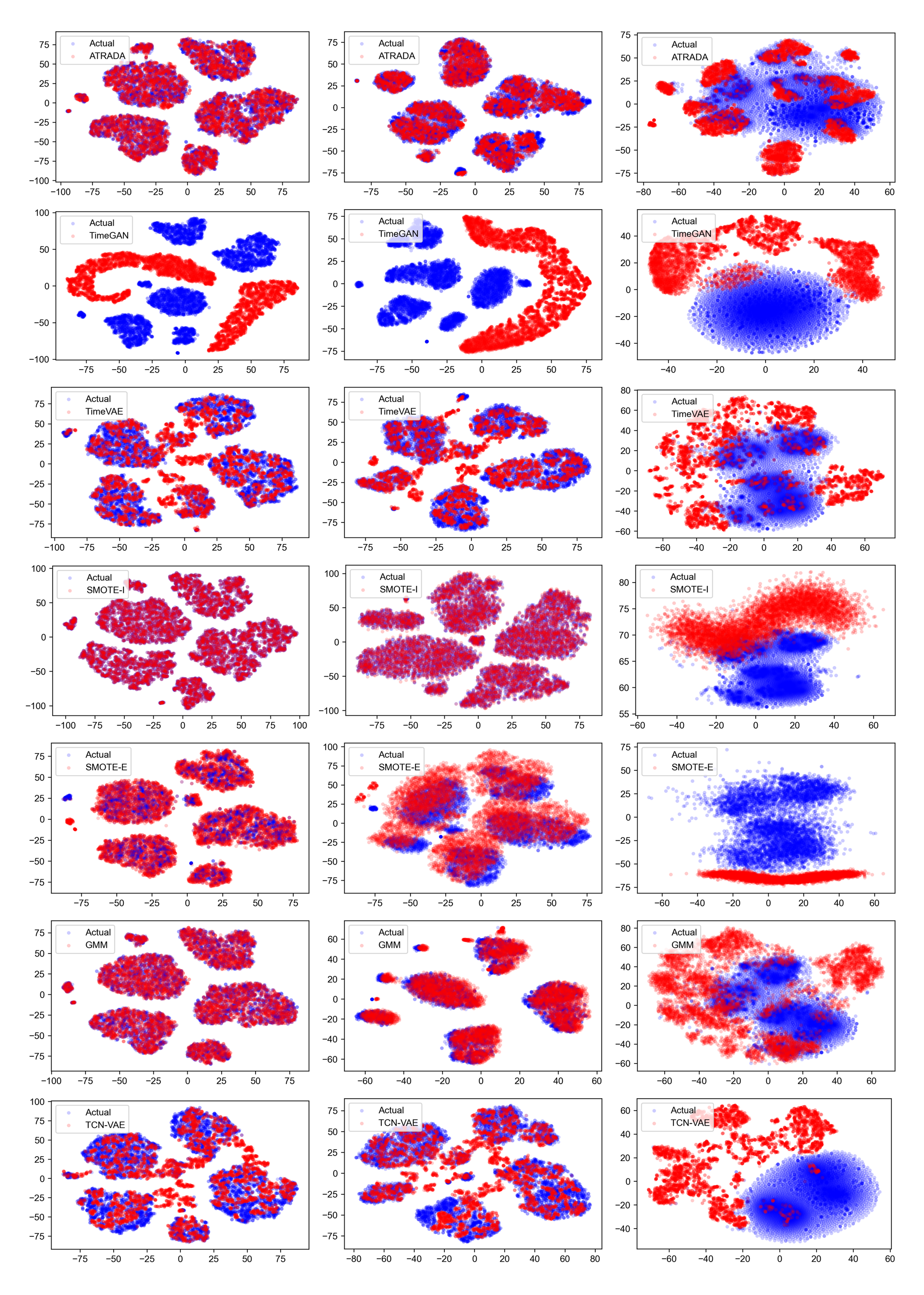}
    \caption{The t-SNE visualizations for ATRADA and all baselines. Each column denotes the generated trajectories' position ($1^{\text{st}}$ column), velocity ($2^{\text{nd}}$ column), acceleration ($3^{\text{rd}}$ column), respectively. Each row depicts the t-SNE visualizations for each method in the following order: (1) ATRADA, (2) TimeGAN, (3) TimeVAE, (4) SMOTE-I, (5) SMOTE-E, (6) GMM, and (7) TCN-VAE. Blue indicates the actual trajectories, while red indicates the generated trajectories.}
    \label{t-sne}
\end{figure*}

\begin{figure*}[t!]			
	\centering
	\includegraphics[width=0.8\textwidth]{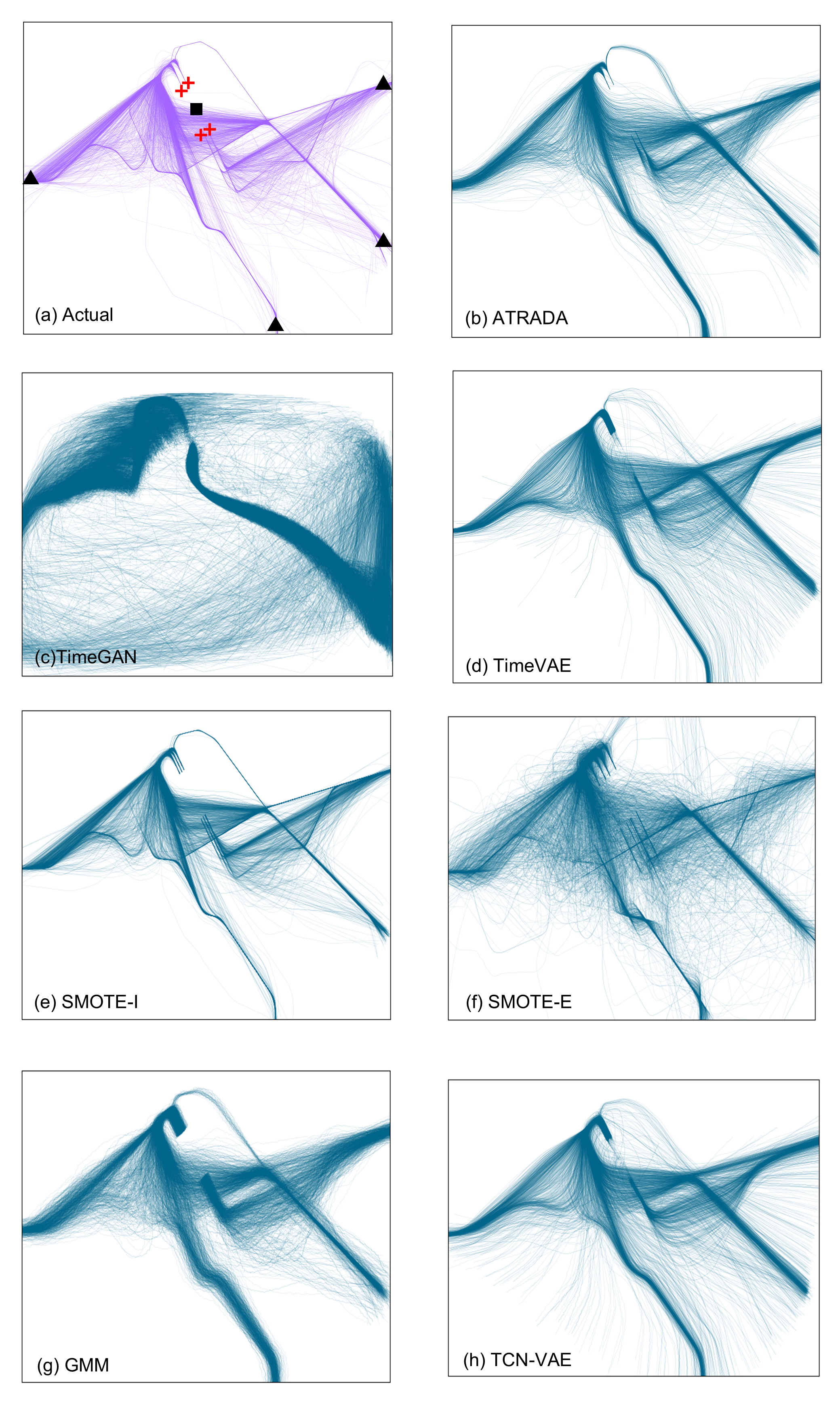}
    \caption{Top view of generated trajectories by ATRADA and all other baselines. The entry points into the airspace and the airport marked by black triangles and squares, respectively. The FAFs for the runway are shown by red "+" symbols.}
    \label{last_trj}
\end{figure*}

The generated trajectories from all baselines are additionally visualized using t-SNE and 2D plots, as depicted in Figure~\ref{t-sne} and Figure~\ref{last_trj}.

\end{appendices}

\clearpage

\section*{Conflict-of-Interest Statement}
On behalf of all authors, the corresponding author states that there is no conflict of interest.

\bibliography{sn-bibliography}

\end{document}